\documentclass[extended abstractpaper, 10pt, conference]{ieeeconf}
\IEEEoverridecommandlockouts
\usepackage{cite}
\usepackage{amsmath,amssymb,amsfonts}
\usepackage{algorithmic}
\usepackage{graphicx}
\usepackage{textcomp}
\usepackage{xcolor}
\def\BibTeX{{\rm B\kern-.05em{\sc i\kern-.025em b}\kern-.08em
    T\kern-.1667em\lower.7ex\hbox{E}\kern-.125emX}}
\usepackage{flushend}

\begin{document}
\newcommand{\ab}[1]{\textcolor{red}{#1}}
\newcommand{\comm}[1]{\textcolor{blue}{#1}}

\title{Creating a Segmented Pointcloud of Grapevines by Combining Multiple Viewpoints Through Visual Odometry}

\author{
Michael Adlerstein$^{1}$, 
Angelo Bratta$^{1}$,
Jo\~ao Carlos Virgolino Soares$^{1}$,
Giovanni Dessy$^{1}$, \\
Miguel Fernandes$^{1}$, 
Matteo Gatti$^{2}$,
Claudio Semini$^{1}$ 
\thanks{$^{1}$ Dynamic Legged Systems Laboratory, Istituto Italiano di Tecnologia (IIT), Genova, Italy.
Email: {\tt\small name.surname@iit.it}}
\thanks{$^{2}$ Department of Sustainable Crop Production (DI.PRO.VE.S.), Università Cattolica del Sacro Cuore, Via Emilia Parmense 84, 29122 Piacenza, Italy.}
}

\maketitle

\begin{abstract}
Grapevine winter pruning is a labor-intensive and repetitive process that significantly influences the quality and quantity of the grape harvest and produced wine of the following season. It requires a careful and expert detection of the point to be cut.
Because of its complexity, repetitive nature and time constraint, the task requires skilled labor that needs to be trained. This extended abstract presents the computer vision pipeline employed in project Vinum, using detectron2 as a segmentation network and keypoint visual odometry to merge different observation into a single pointcloud used to make informed pruning decisions.
\end{abstract}

\section{Introduction}
Winter pruning is a critical, labor-intensive task in vineyards, crucial for balancing yield and grape quality to maximize income. It requires 80 to 120 man-hours per hectare annually. The process becomes even more significant considering labor shortages in agriculture. Skilled workers in vineyards must know plant anatomy to make the correct pruning cuts.
Identifying the correct pruning regions is a non-trivial task that requires a trained eye in order to distinguish between the different organs, creating a need for a reliable and accurate system to identify such regions and pruning points.  With the rapid advancements in machine learning and neural networks, autonomous methods have been explored to bridge this gap. 

In the agricultural sector neural networks have been used for weeds detection \cite{di2017automatic} \cite{milioto2018real} and for fruit detection and segmentation, as demonstrated by Borianne et al.~\cite{borianne2019deep}, and Santos et al.~\cite{santos2020grape}. 
Further applications include plant phenotyping where Grimm et al.~\cite{grimm2018adaptive} utilized neural networks for detailed plant analysis.

In the context of vineyard pruning, Gentilhomme et al. \cite{gentilhomme2023towards} presented ViNet, a network specifically designed to identify nodes within a grapevine plant. Their method relies on a stacked hourglass network \cite{newell2016stacked} for detailed reconstruction. The paper also introduces the 3D2Cut dataset, which provides annotated data with node information against synthetic backgrounds. 
Guadagna et. al. ~\cite{guadagna2023using} demonstrated that a  R-CNN model is successful in achieving organ segmentation (cordon, arm, spur, cane and node are the visible organs on a grapevine) demonstrating promising results in terms of recall, highlighting a detection performance for visible co-planar simple spurs with high precision and recall.
Notable contributions also came from Fernandes et al.~\cite{fernandes21} which demonstrated that it is possible to detect the potential pruning points based on 3D scans of the pruning regions, and Botterill et al.\cite{botterill2017robot}  and Silwal et al. \cite{silwal2021bumblebee} which developed robot prototypes towards the automation of pruning.

Despite the promising advancements showcased, neural networks are not always able to give a one-shot complete segmentation of the plant. This is because the dataset is not diverse enough and the labelling of the different organs is a non-trivial task that requires a specialized person. In addition, some organs might be occluded from a certain point of view, requiring more than one image; as a result the neural networks are sometimes unable to give an accurate result from just one image. This extended abstract builds upon the previous work done inside the  Vinum project\footnote{https://vinum-robot.eu/} (e.g. \cite{guadagna2023using}, \cite{fernandes21}). In particular, we describe here an approach for 2-dimensional visual odometry used to combine different viewpoints into one segmented pointcloud, resulting in a detailed 4D structure (x y z position and class label), which can be used by a robot to accurately prune the grapevines. 

\section{Method}
\begin{figure}[h] 
\centering
    \includegraphics[width=0.47\textwidth]{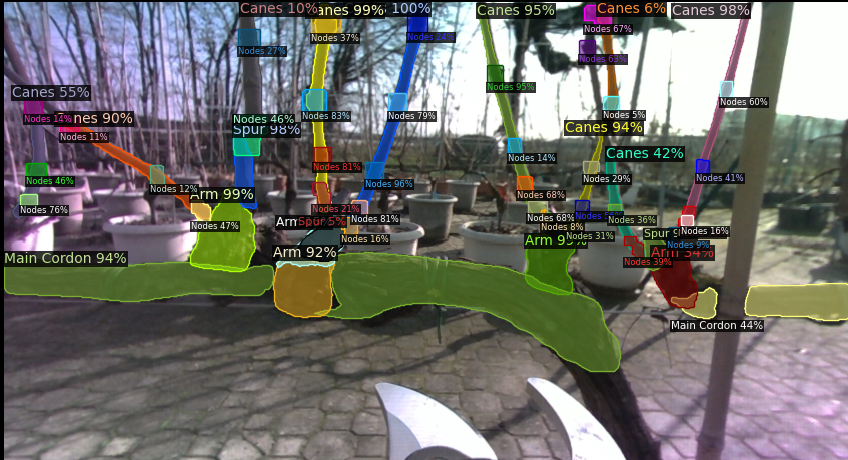}
      \caption{Output inference of detectron2, giving precise segmentation labels of the five classes.}
      \label{fig:segmentation_img}
\end{figure}

A convolutional neural network using the detectron2 backbone is trained on a dataset containing segmented images of grapevines, enabling the detection of the five main organ classes (cordon, arm, spur, cane and node) \cite{guadagna2023using}. This training allows for precise inferences on new images, as illustrated in Fig. \ref{fig:segmentation_img}, showcasing a recall of 81\% and precision of 97\%.
To construct a more accurate representation of the plant, pictures from different viewpoints are taken using a robotic manipulator equipped with an RGB-D camera, covering the entire plant. This process ensures that the same part of the plant appears in multiple frames, facilitating the extraction of keypoints. The keypoints are then matched across different instances and projected into 3D space. Following this, a transformation matrix is calculated, enabling the transformation of the entire pointcloud into the same frame of reference as the robot, thereby merging multiple distinct viewpoints into a single pointcloud.

\subsection{Keypoint Matching}
\begin{figure}[h]
  \centering

    \centering    \includegraphics[width=0.47\textwidth]{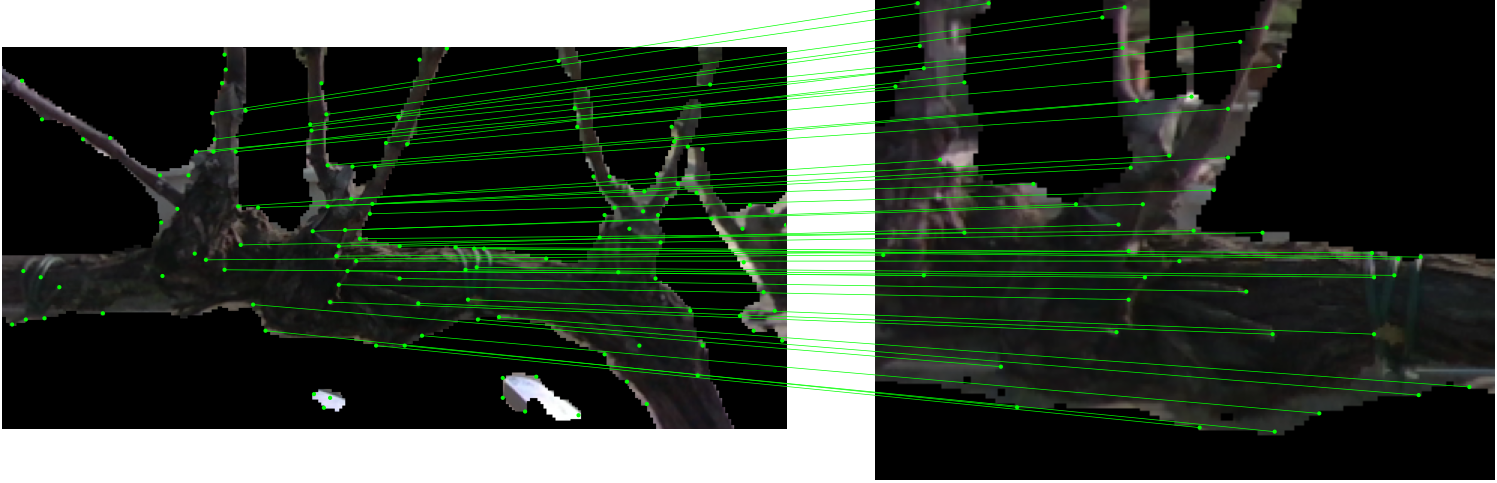}
    \caption{Keypoint detection and matching using Superpoint and Lightglue.}
    \label{fig:key_point}
\end{figure}

Traditional descriptor methods, such as ORB \cite{rublee2011orb}, were not suitable for the task due to the homogeneous colour and texture of the plant. Advancements in the state of the art have led to the adoption of SuperPoint \cite{detone2018superpoint}, which offers a balanced approach between detecting a substantial number of keypoints and maintaining manageable inference times. LightGlue \cite{lindenberger2023lightglue} was chosen as the matcher due to its speed and matching reliability. 
These novel approaches, utilizing neural networks and attention, represent a significant improvement over traditional methods by offering a superior trade-off between the speed of inference and the number of detected features. 
The improvements in the detection and the matching allow Visual odometry to merge pointclouds, which are spatially distant from each other reducing the number of frames used whilst also maximising the amount of new information detected, as shown in Fig. \ref{fig:key_point}. 

\subsection{3D Projection and Registration}
Once the keypoints are matched, the registration is achieved through the minimization in the Orthogonal Procrustes problem. This method seeks to optimally map a set of keypoints between two distinct frames of reference. Allowing for the alignment of two different pointclouds into the same frame.
Mathematically, the problem is formulated as follows:
\begin{equation}
\min_{\mathbf{Q}} \| \mathbf{A} - \mathbf{B}\mathbf{Q} \|_{F} \quad \text{subject to} \quad \mathbf{Q}^T\mathbf{Q} = \mathbf{Q}\mathbf{Q}^T = I,
\end{equation}
where $\mathbf{A}$ and $\mathbf{B}$ are vectors containing the positions of the keypoints to be matched. The term $\| \cdot \|_{F}$ denotes the Frobenius norm, and $\mathbf{Q}$ represents the rotation matrix that minimizes the equation.

To determine the quality of alignments, two metrics are used: RMSE (Root Mean Square Error) and fitness. If the fitness value is too low or the RMSE value is too high, the match is discarded, and a new set of frames is selected for alignment. After aligning all pointclouds to the same frame of reference (the camera frame), a final transformation then merges pointclouds into the base frame, in order for the pointcloud to be in the same frame as the robotic manipulator.

For global optimization, pose graph optimization is employed to refine locally optimized transformations from the final matched pointclouds. This approach uses a "pose graph" structure to achieve optimal overlap between pointclouds. The pose graph comprises nodes representing the positions of the pointclouds and edges that represent transformations between these nodes. 

\subsection{Clustering}
\begin{figure}[h]
    \centering
    \includegraphics[width=0.48\textwidth]{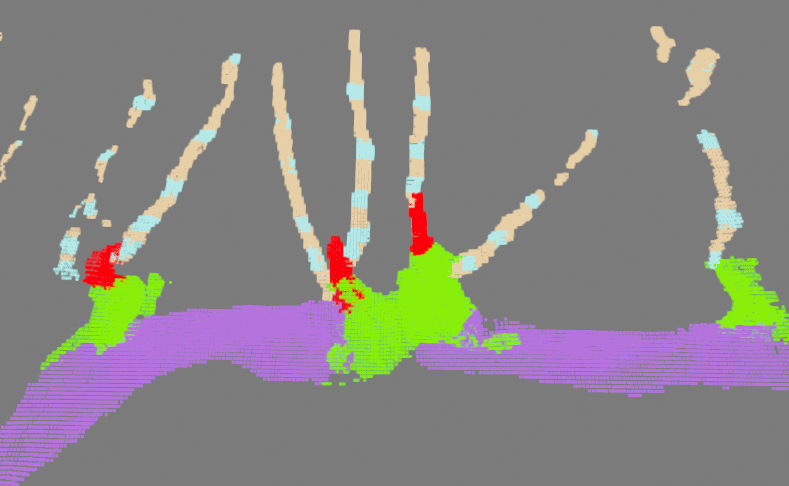}
    \caption{Final pointcloud with segmented organs.}
    \label{fig:segmented_pc}
\end{figure}

Once the entire scan set of RGB-D images is obtained and correctly aligned, they are merged into a single pointcloud representing the entire structure of the plant. 
HDBSCAN~\cite{stewart2022implementation} is used to address the clustering issue, as it works well with different densities of points allowing for the creation of comprehensive clusters.
The process of clustering points is crucial for distinguishing between different instances of the same object type, such as individual organs in a plant as shown in Fig. \ref{fig:segmented_pc}. This enables detailed analysis of the plant, which is essential for making informed decisions regarding which parts to keep or remove.
The HDBSCAN merges all the points belonging to the same class from different scans, discarding duplicate points and selecting the instance which has the highest probability.
 
\section{Conclusion and Future Work}
This extended abstract presented a novel approach for scanning grapevines, allowing the creation of an accurate pointcloud representation that can be used by a robotic manipulator to plan for accurate pruning. This approach is able to produce a 3D representation of the plant and its segmented organs in matter of minutes. The merging of pointclouds from several frames enables the system to account for neural network failures, making the system more robust. 
The system occasionally fails to cluster regions depending on the morphology of the plant due to the nature of HDBSCAN. Furthermore, due to the resolution of the camera some ghosting can be observed on the canes resulting in a less precise pruning region calculation. To account for these shortcomings, more work has to be done to have a better clustering algorithm.

\bibliography{root}

\end{document}